\documentclass{article}
\usepackage[accepted]{mlsys2024}
\usepackage{microtype}
\usepackage{graphicx}
\usepackage{subfigure}
\usepackage{booktabs} 
\usepackage{tabularx}
\usepackage{enumitem}
\usepackage{pifont}
\usepackage{xspace}
\usepackage{color, colortbl}
\usepackage{url}

\usepackage[breaklinks]{hyperref}
\usepackage{breakurl}
\usepackage{tikz}
\usepackage{amsmath}
\usepackage[textsize=tiny,textwidth=0.6in]{todonotes}
\usepackage{setspace}
\usepackage{titlecaps}
\usepackage{listings}

\def\compactify{\itemsep=2pt \topsep=2pt \partopsep=1pt \parsep=1pt \leftmargin=1.6em}
\let\latexusecounter=\usecounter

\newcommand{\towermodule}{TM\xspace}
\newcommand{\dmtembed}{TP\xspace}
\newcommand{\lrm}{XLRM\xspace}
\newcommand{\spt}{semantic-preserving tower transform\xspace}
\newcommand{\sptf}{SPTT\xspace}
\newcommand{\strongbaseline}{Strong Baseline\xspace}

\def\compactify{\itemsep=2pt \topsep=2pt \partopsep=1pt \parsep=1pt \leftmargin=1em}
\setlist{nolistsep}

\mlsystitlerunning{Disaggregated Multi-Tower: Topology-aware Modeling Technique for Efficient Large Scale Recommendation}

\begin{document}

\twocolumn[
\mlsystitle{Disaggregated Multi-Tower: Toplogy-aware Modeling Technique for Efficient Large Scale Recommendation}


\mlsyssetsymbol{equal}{*}

\begin{mlsysauthorlist}
\mlsysauthor{Liang Luo}{meta}
\mlsysauthor{Buyun Zhang}{meta}
\mlsysauthor{Michael Tsang}{meta} \\
\mlsysauthor{Yinbin Ma}{meta}
\mlsysauthor{Yuxin Chen}{meta}
\mlsysauthor{Ching-Hsiang Chu}{meta}
\mlsysauthor{Yanli Zhao}{meta}
\mlsysauthor{Shen Li}{meta} \\

\mlsysauthor{Yuchen Hao}{meta}
\mlsysauthor{Guna Lakshminarayanan}{meta}\\
\mlsysauthor{Ellie Dingqiao Wen}{meta}
\mlsysauthor{Jongsoo Park}{meta}
\mlsysauthor{Dheevatsa Mudigere}{nvidia}
\mlsysauthor{Maxim Naumov}{meta} \\

\end{mlsysauthorlist}

\mlsysaffiliation{meta}{Meta AI}
\mlsysaffiliation{nvidia}{Nvidia, work done while at Meta}
\mlsyscorrespondingauthor{Liang Luo}{liangluo@meta.com}

\mlsyskeywords{Machine Learning, MLSys}
\vskip 0.3in

\begin{abstract}
We study a \textit{mismatch} between the deep learning recommendation models' flat architecture, common distributed training paradigm and hierarchical data center topology. To address the associated inefficiencies, we propose Disaggregated Multi-Tower (DMT), a modeling technique that consists of (1) Semantic-preserving Tower Transform (\sptf), a novel training paradigm that decomposes the monolithic global embedding lookup process into disjoint towers to exploit data center locality; (2) Tower Module (\towermodule), a synergistic dense component attached to each tower to reduce model complexity and communication volume through hierarchical feature interaction; and (3) Tower Partitioner (\dmtembed), a feature partitioner to systematically create towers with meaningful feature interactions and load balanced assignments to preserve model quality and training throughput via learned embeddings. We show that DMT can achieve up to 1.9$\times$ speedup compared to the state-of-the-art baselines without losing accuracy across multiple generations of hardware at large data center scales.
\end{abstract}
]



\printAffiliationsAndNotice{}  
\section{Introduction}
Recommendation models have played a critical role in online services including search engines, social media, and content platforms. Recent advancements in recommendation models are in large part brought by the use of neural networks and the exponential growth in model capacity. To date, deep-learning powered recommendation models with billions to even of trillions of parameters are no longer uncommon~\cite{aibox, jiang2019xdl, lian2021persia, zhao2022communication}. 


Modern recommendation models consist of a \textit{sparse} component that converts categorical features into dense representations using \textit{embedding lookup tables}, and a \textit{dense} neural network component that interacts the embedding of categorical features and dense features and produces an output. Since the embedding tables can be huge, the state-of-the-art practices train these models in a hybrid fashion: the sparse part is trained in a model parallel and the dense part in a data parallel manner~\cite{naumov2020deep, mudigere2021high, lian2021persia, dhen}, and the embedding lookup requests/responses are routed to respective accelerators via AlltoAll collectives and dense parameters are synchronized through AllReduce operations.  


\begin{figure}
\begin{center}
	\centering
	\footnotesize
	\includegraphics[width=\columnwidth]{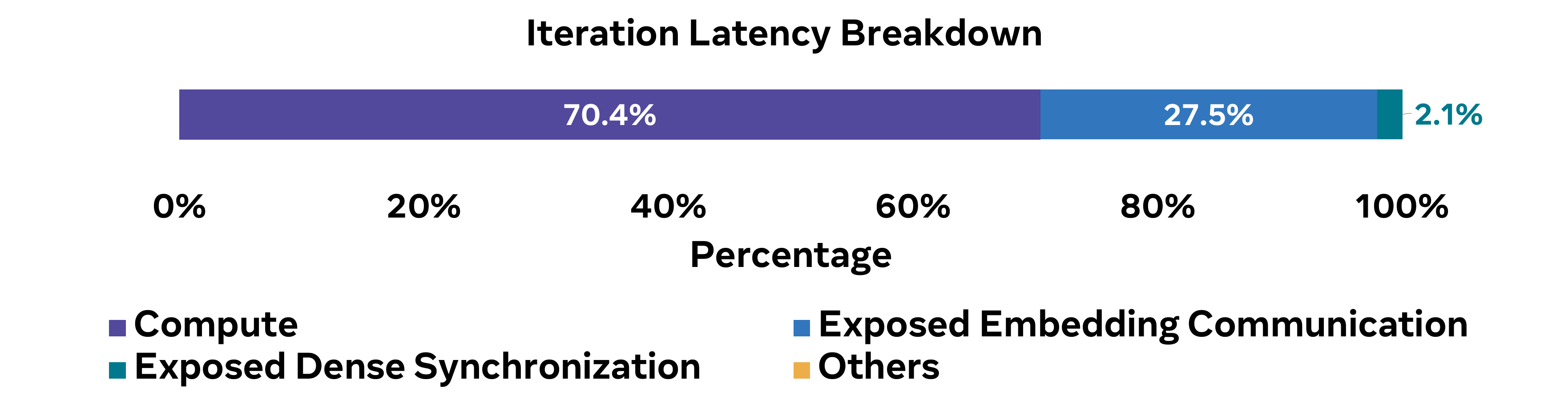}
\end{center}
\caption{Exposed latency breakdown of training DCN on a cluster with 64 H100 GPUs with state-of-the-art optimizations.}
\label{fig:exposed_latency_dlrm}
\end{figure}

\begin{figure}[!t]
\begin{center}
	\centering
	\footnotesize
	\includegraphics[width=\columnwidth]{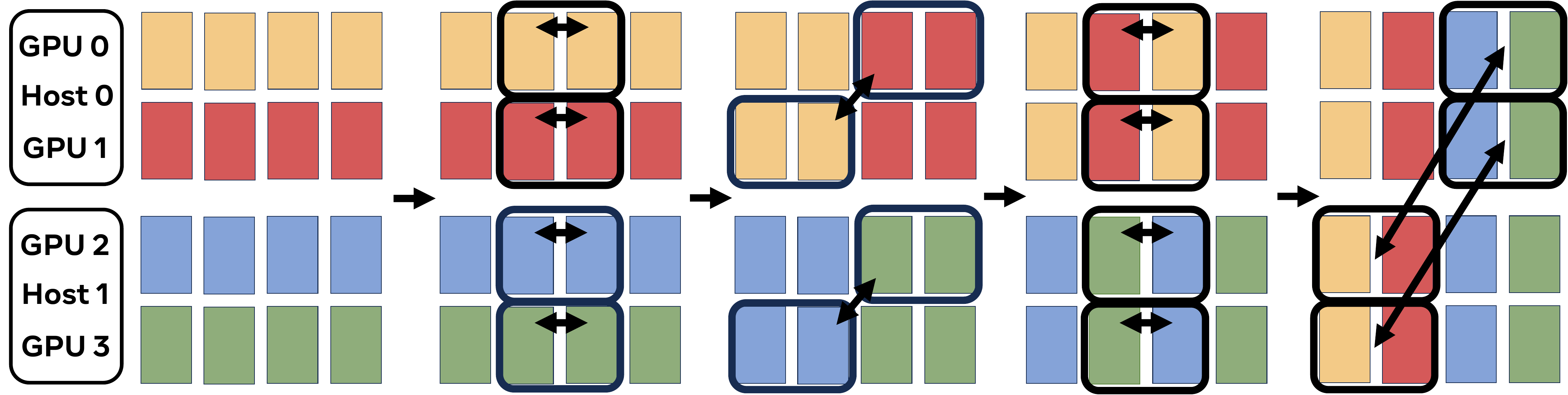}
\end{center}
\caption{An example of a topology-aware communication pattern that is semantically-equivalent to AlltoAll.} 
\label{fig:banner}
\end{figure}

This training paradigm has faced increasingly more challenges due to the divergence of two model and hardware scaling trends (\S\ref{sec:background}): (1) the use of more sparse features and higher dimensional embeddings~\cite{kang2020learning} increases the communication volumes of the AlltoAll communications; (2) the speed of improvements of compute capacity in modern accelerators have significantly outpaced that of the physical network in datacenter environments~\cite{luo2018parameter, sapio2019scaling}. For example, Table~\ref{table:dlrm_scaling_trend} lists recent datacenter developments, where the floating point compute power improved by 60$\times$, whereas the scale-out bandwidth only increased by 4$\times$ within the same time frame. 
Consequently, the bottleneck of recommendation model training has shifted to the communication of the embedding lookup request and response and synchronization of the dense gradients. Even with modern optimizations such as quantized communications~\cite{yang2020training}, automatic column-wise sharding~\cite{zha2023pre}, overlapped compute and communication~\cite{9499904} and fused embedding lookup kernel~\cite{fbgemm}, in a typical datacenter environment with 64 GPUs and fast RDMA network, up to 30\% of time within an iteration is spent explicitly waiting on the network communication, see Figure~\ref{fig:exposed_latency_dlrm}, causing a large inefficiency in the datacenter utilization.

Recent proposals including Piper~\cite{tarnawski2021piper}, Alpa~\cite{zheng2022alpa}, Megatron~\cite{narayanan2021efficient}, and ZeRo~\cite{rasley2020deepspeed,zhao2023pytorch,zhang2022mics} aim to find alternative paradigms for large scale training. As we show later, they are unable to suggest better parallelism that beats hybrid parallelism for modern recommendation models empirically. This implies that the current hybrid parallelism represents a near-optimal configuration in the known parallelism search space, and the fact that it fails to achieve satisfactory efficiency calls for more efforts in the domain of improving recommendation model training performance.

Fundamentally, we show that this inefficiency stems from a mismatch between the model architecture, training paradigm and datacenter topology: typical datacenters have different hierarchies of communication bandwidth, with drastically faster intra-host (e.g., NVLink) than cross-host connections (e.g., RDMA), but most recommendation models are flat, meaning both the embedding distribution and the feature interaction processes are global and monolithic.

\begin{table}[!t]
	\centering
	\resizebox{\columnwidth}{!}
	{
    	\begin{tabular}{|c|c|c|c|}
    	\hline
    	    System           & Peak FP Perf  & Scale-out/GPU  & Scale-up/GPU (unidirection) \\
    	    \hline
    	    V100, 2019       &  15.7~TF/s       & 100~Gbps           & 150~GB/s \\
    	    \hline
    	    A100, 2022     &  156~TF/s        & 200~Gbps           & 300~GB/s \\
                \hline
                H100, 2023    &  989~TF/s        & 400~Gbps           & 450~GB/s \\
    	\hline
    	\end{tabular}
	}
	\caption{Recent generational upgrades reported by sources~\cite{8327042,8875650,mudigere2021high,MetasGr63:online,TheIronT92:online} and manufacturer show the improvements of compute capacity significantly outpaces that of the network bandwidth's, making communication the bottleneck.}
    \label{table:dlrm_scaling_trend}
\end{table}

To address this mismatch, we propose the Disaggregated Multi-Tower (DMT) modeling technique that transforms the model and takes advantage of the heterogeneity in datacenter for efficiency improvements (\S\ref{sec:design}). We show that the AlltoAll communication corresponding to a large exposed latency in Figure~\ref{fig:exposed_latency_dlrm} can be efficiently expressed through a set of local shuffles and localized communications shown in Figure~\ref{fig:banner}. Further, we show that through careful design of modules, the latency and communication volume can  be reduced without hurting model quality. Specifically, DMT achieves this by (1) partitioning the sparse features into semantic-preserving and balanced \textit{towers} through \textit{Tower Partitioner} (\dmtembed), a learned feature partitioner; (2) decomposing the global embedding distribution into disjoint towers to exploit datacenter locality, through a novel training paradigm called \textit{\spt} (\sptf); (3) introducing \textit{tower modules} (\towermodule) that are synergistic to \sptf and leverages hierarchical feature interaction to reduce both model complexity and communication volume.

We optimized the implementation of DMT (\S\ref{sec:implementation}) and applied it to two popular and widely-used open source models DLRM~\cite{naumov2019deep} and DCN~\cite{wang2021dcn}) as well as an internal workload. In our experiments we show that across three generations of hardware (Nvidia V100, A100 and H100) on up to 512 GPUs, DMT outperforms the state-of-the-art baseline in training throughput by up to 1.9$\times$ without losing the statistical quality (\S\ref{sec:evaluation}).

\section{Challenges of Large-Scale Recommendation Models Training}
\label{sec:background}
This section provides contexts for the challenges of training large-scale recommendation models in a data center setting. 

\subsection{Deep Learning-based Recommendation Models}
\begin{figure}[!t]
\begin{center}
	\centering
	\footnotesize
	\includegraphics[width=.9\linewidth]{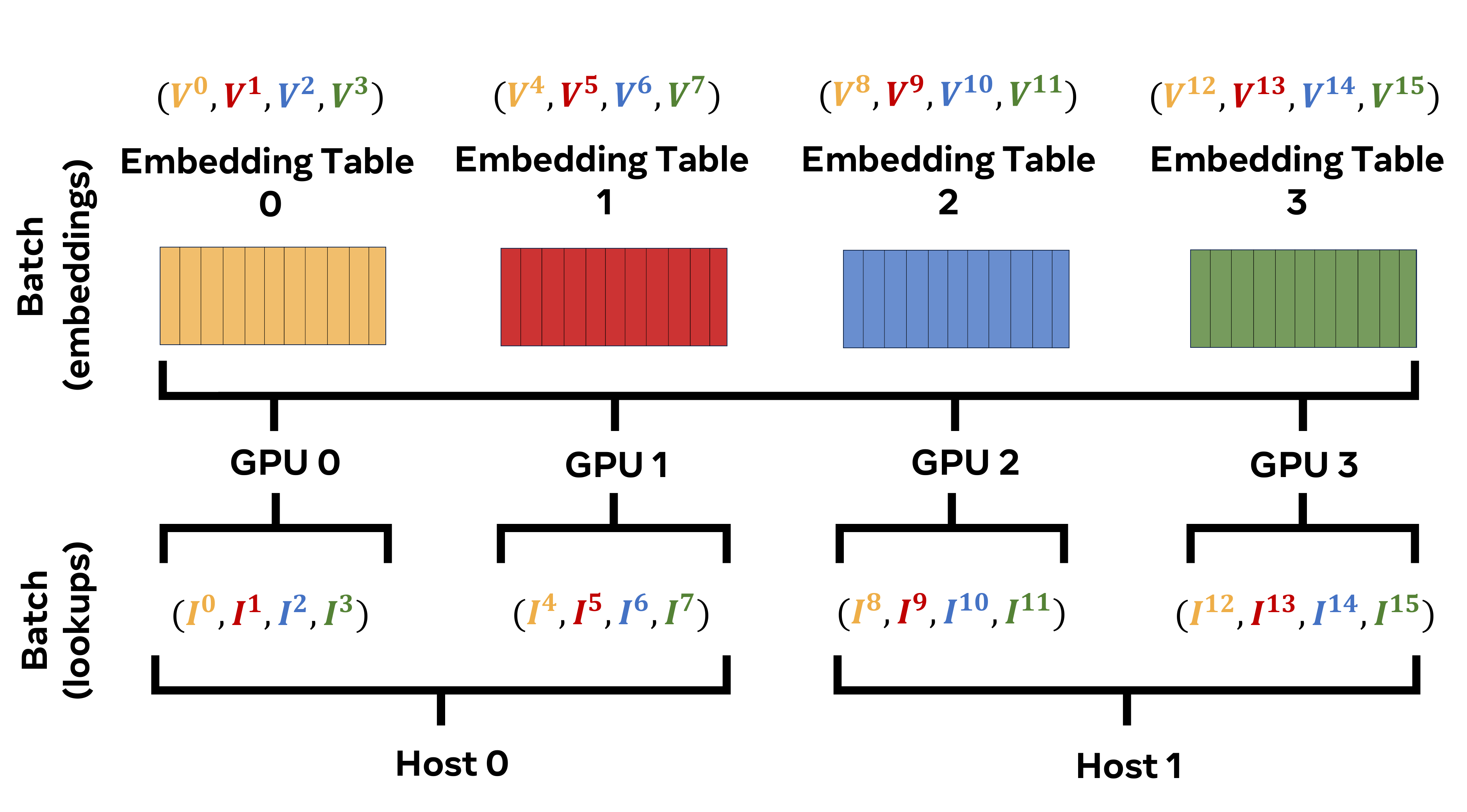}
\end{center}
    \caption{A conceptual view of the embedding lookup process.}
\label{fig:embedding_overview}
\end{figure}

Modern deep learning-based recommendation models share a similar structure: they usually consists of large embedding tables and dense networks~\cite{aibox, jiang2019xdl, lian2021persia, zhao2022communication,dhen}. These models take in both dense (continuous) and categorical (sparse) features as inputs. The sparse features are first processed by the embedding tables, from which embeddings are acquired. The dense features are then joined with the embeddings for feature interaction. Finally, the outcome of this interaction is fed to another dense network where a prediction is produced. Perhaps the most unique aspect about recommendation model is the embedding lookup process, and we explain it more in Figure~\ref{fig:embedding_overview} using an example: we have 4 GPUs spanning over 2 hosts, and each of which is given an input sample $(I)$. Each sample contains 4 color-coded features, with inputs to these features labeled $I^{4r}...I^{4r+3}$ for clarity of explanation, where $r$ is the rank of the GPU. Each $I^k$ is itself a tuple of discrete values, and the goal of embedding lookup is to convert each $I^k$ into a dense representation $V^k$ for all $k$s. 




\subsection{Training Process}

\begin{figure}[!t]
\begin{center}
	\centering
	\footnotesize
	\includegraphics[width=\columnwidth]{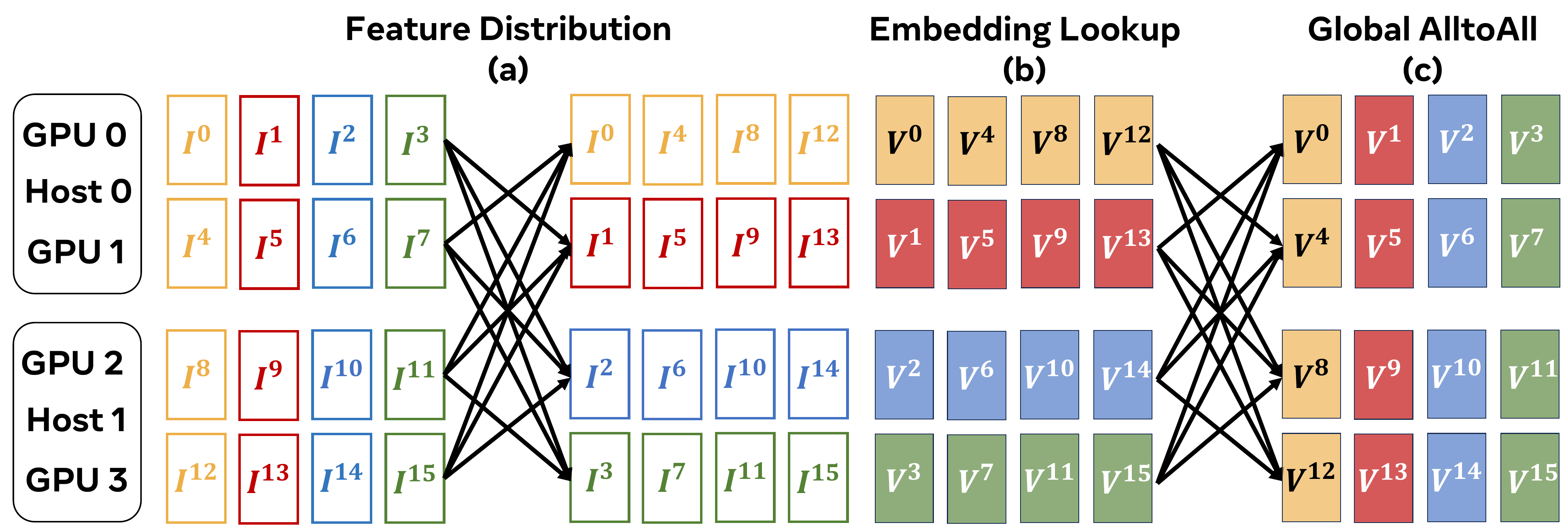}
\end{center}
\vspace*{-3mm}
\caption{An operational view of the classic embedding lookup process in current systems.} 
\label{fig:hybrid_training}
\end{figure}

The current state-of-the-art frameworks, Neo~\cite{mudigere2021high} and TorchRec~\cite{torchrec}, train these models using a hybrid parallelism: the embedding tables are sharded to different devices in model parallelism, and the dense networks run as data parallel. The adoption of the hybrid training paradigm is a result of two factors: (1) embedding tables are simply too large to be synchronized as a dense component and (2) in each iteration, access to these tables is sparse so only a few entries are updated. Since the training process of dense components is similar to models in other domains, we focus our discussion on the embedding lookup mechanism, highlighted in Figure~\ref{fig:hybrid_training}.



The setup is the same as Figure~\ref{fig:embedding_overview} to facilitate illustration. At first, the local sparse features are communicated with devices where their respective embedding tables reside through a global AlltoAll communication (step a). Each GPU is then tasked with looking up embeddings for the global batch for the categorical features assigned to it (step b). The lookup process involves parallel queries into embedding table entries and pooling if necessary. Afterwards, another AlltoAll communication is used to return embeddings to each GPU (step c). This concludes the training forward pass for embedding lookup. In the backward pass, gradients to activated embedding table entries are routed through another AlltoAll.
 
 



\subsection{Training Bottleneck Analysis}
This training paradigm incurs significant overheads and hinders efficient training of such models in a typical datacenter environment, with the main bottleneck being the inefficient AlltoAll collectives triggered by the global embedding distribution process, as shown in Figure~\ref{fig:exposed_latency_dlrm}.

\begin{figure}[!t]
\begin{center}
	\centering
	\footnotesize
	\includegraphics[width=\columnwidth]{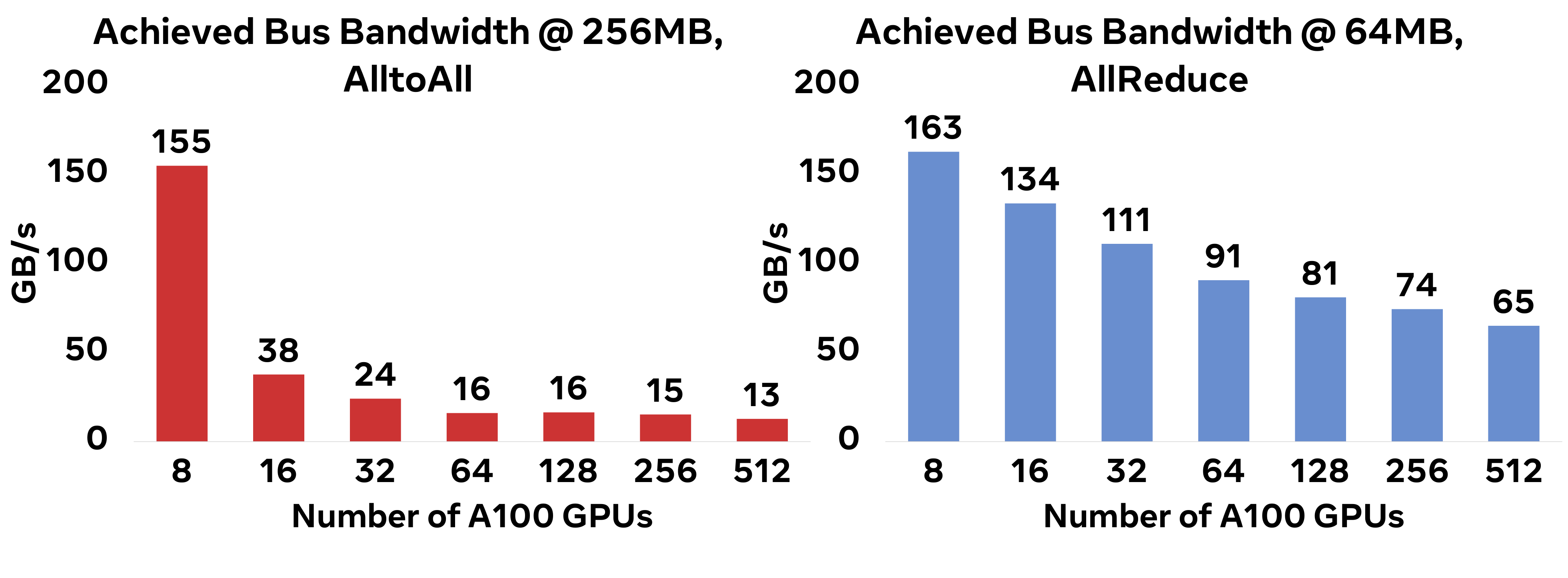}
\end{center}
\vspace*{-3mm}
\caption{Scalability (weak scaling) of NCCL collectives in a datacenter environment, 8 GPUs/nodes, measured in bus bandwidth.}
\label{fig:collective_perf}
\end{figure}

\subsubsection{Degrading Collective Performance vs Scale.} 
\label{sec:collectives_perf}
The direct reason for poor training efficiency is the subpar collective performance. To show this, we measure the bus bandwidth achieved running DLRM's collectives, Allreduce and AlltoAll with NCCL (v2.18.3), using typical buffer size close to the one used in typical DLRM training rounded up to the nearest power of 2 (64MB of dense size, and 256MB of embedding size at the batch size of 16K, in float). 

As shown in Figure~\ref{fig:collective_perf}, performance of collectives drops sharply versus scale, and the consequence of this is exacerbated by the fact that AlltoAll and AllReduce collectives are invoked at least 3 and 1 time(s) per iteration, respectively, in hybrid parallelism.

\subsubsection{Mismatched model architecture, training paradigm, and data center topology.}
\label{sec:mismatch_analysis}

The degrading communication performance is hardly a surprise at a large scale~\cite{liu2017incbricks,9308656, 7830486, luo2020plink, SRIFTY}, therefore it is superficial to blame the poor efficiency solely on it. Here, we would like to argue that the fundamental issue is the mismatch between model architecture, training paradigm and data center topology:

\begin{itemize}[leftmargin=*,noitemsep,topsep=0pt]
    \item The model architecture, which usually requires global feature interaction, together with the training paradigm, dictate multiple rounds of global AlltoAlls for embedding lookup. This means we need to send a byte on wire (forward and backward pass) just to read/write a byte (times pooling factor) from the GPU memory during embedding lookup (assuming no quantization). Since the discrepancy between memory bandwidth and network bandwidth is huge (orders of magnitudes) in current systems, this makes network a bottleneck in training.
    
    \item The training paradigm fails to exploit locality within a single host: the fast NVLink connection within a single host delivers much higher bandwidth compared to cross-host links. Since the latency of AlltoAll is bounded by the slower cross-host RDMA links, NVLink offers limited benefits for the final collective performance.
\end{itemize}

\subsection{Ineffectiveness of Existing Solutions}
Given the above analysis, it is worth asking: (1) can recent sharding improvements solve the problem? and (2) can we improve performance by only optimizing the training paradigm without modifying the model itself? Unfortunately, the answer to both questions is likely, no. 

On one hand, NeuroShard~\cite{zha2023pre} improves embedding distribution process with better load balance, but it cannot solve the problem of high AlltoAll latency even with a perfect load balance, as we show in \S\ref{sec:evaluation}. On the other hand, as we show later, applying recent advancements including Piper, Alpa, Megatron, and ZeRo on recommendation models have barely any effect, as they cannot address the nontrivial overhead from sparse components in recommendation models.


\begin{figure}[!t]
\begin{center}
	\centering
	\footnotesize
	\includegraphics[width=.9\columnwidth]{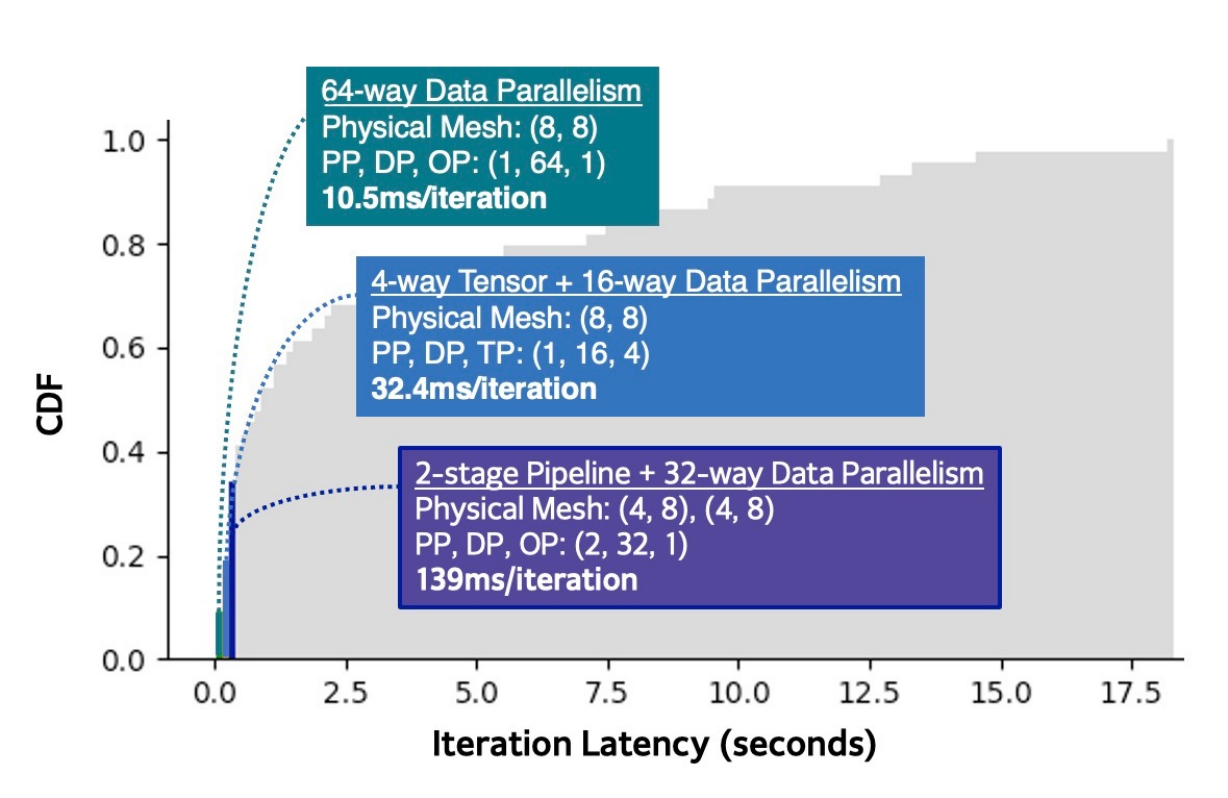}
\end{center}
\vspace*{-4mm}
\caption{CDF of the iteration latency for various parallelism configurations running DLRM with Alpa.}
\label{fig:alpa_dlrm}
\end{figure}

For example, we leveraged Alpa to search for the optimal parallelism strategy for the dense part of DLRM on 64 A100 GPUs by enumerating various physical and logical meshes, with the results summarized in Figure~\ref{fig:alpa_dlrm} as a CDF of the iteration latency for each valid configuration. We then highlight the bins where the fastest configuration that uses data, tensor and pipeline parallelism lies. Evidently, data parallelism stands out alone as the fastest parallelism for the dense part of DLRM. As a result, the widely adopted hybrid training approach indeed represents a (near) optimal configuration in the known parallelism search space. We cannot address the fundamental problems with existing tools. 

\section{Disaggregated Multi-Tower}
\label{sec:design}
\begin{figure*}[!t]
\begin{center}
	\centering
	\footnotesize
	\includegraphics[width=.95\linewidth]{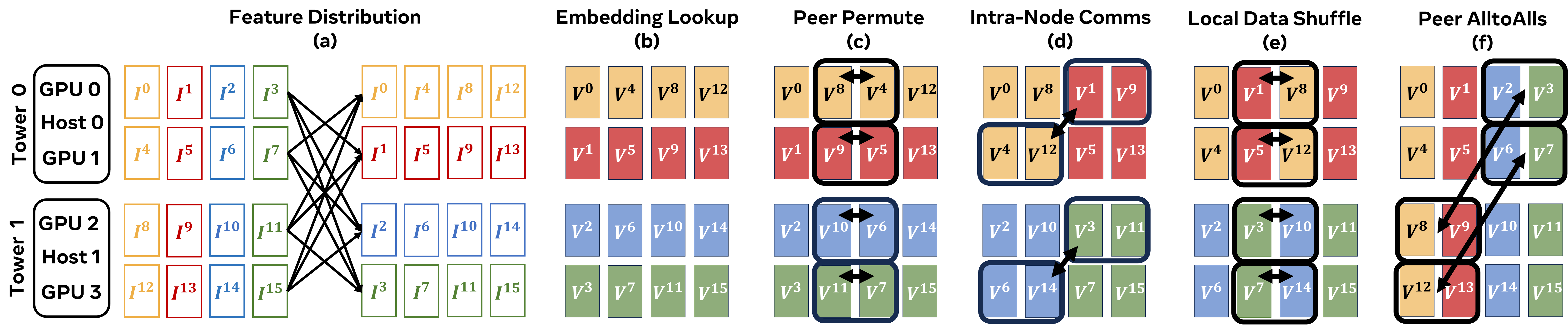}
\end{center}
\vspace*{-3mm}
\caption{An operational view of the topology-aware embedding lookup process using \spt.} 
\label{fig:tower_transform}
\end{figure*}

We now describe DMT, a topology-aware modeling technique designed to address the fundamental mismatch of model, paradigm and data center topology. We introduce 3 main concepts that support DMT: (1) a \spt (\sptf) that exploits data center topology for faster communication; (2) a tower module (\towermodule) synergistic with \sptf for model complexity and communication volume reduction; and (3) a learned, balanced feature partitioner (\dmtembed) that generates effective tower partition.


\subsection{\titlecap{\spt}}
In this paper, we call a group of sparse features, dense layers that consume their embeddings, together with a group of GPUs that host them a \textit{tower}. Typically, a tower is constructed on a collection of accelerators that have high communication locality (e.g., GPUs within a host). Originally, each model can be viewed as a single tower because feature interactions are global. 

In this section, we show that if we have decomposed the single-towered model into a multi-towered variant by partitioning features into different groups, then through the use of \sptf, we can improve its training performance by tapping into data center topology without changing the semantics of the model. Without losing generality, we first assume the features are pre-partitioned for us, and a later section explains how such partitions can be obtained. 


From an operational perspective, \sptf breaks down the global embedding AlltoAll operation into a series of steps, where each step is either a data shuffle operation, an intra-host collective, or a cross-host communication. 

\subsubsection{\sptf: a Walk-through}
To best illustrate the idea of \sptf, we use the same example as in Figure~\ref{fig:hybrid_training}, where we have four features assigned to two towers, and each tower consists of 2 GPUs within the same host. Concretely, tower 0 consists of the orange and red sparse features assigned to host 0, and tower 1 with the blue and green sparse features assigned to host 1. The \sptf process starts with the same global feature distribution via an AlltoAll call (step a), followed by the local embedding lookup process (step b). We then proceed with a \textit{peer permute} to rearrange the received features on each GPU into a \textit{peer order} (step c). Formally, we define peers of a GPU ranked $g_i$ to be the collections of GPUs whose rank $g_j$ satisfies $g_i \% L = g_j \% L$, where $L$ is the total number of GPUs on a host, and the peer order is defined as the sorted total order of all GPUs with rank $g_i$ using the tuple $(g_i \% T, g_i // L)$ as key, where $T$ is the number of towers (hosts). In our example, the peer order of GPUs is (0, 2, 1, 3). 

With the embeddings are now in peer order, step d, an intra-host communication (usually another AlltoAll operation) follows to exchange partial embeddings with local GPUs such that each GPU holds embeddings results for all its peers for all the features (embedding tables) sharded to it (in our case, we have each GPU holding only one feature/table). Although step d is a collectives communication, it is highly efficient thanks to the fast NVLink interconnection.

Step e involves a series of data shuffle operations - we first view the resulting tensor of step d by stacking it into a 2D tensor with dimension (features, peers), then we transpose it to the layout of (peers, features), and finally view it as a flattened tensor (peers $\times$ features) for the final step. The last step (step f) involves concurrent peer AlltoAlls, where each GPU communicates with its peers. With a total number of $G$ GPUs in the cluster, there will be $L$ such AlltoAlls in parallel, with each AlltoAll operate within a world of size equals to the number of towers $T = G // L$. 

\subsubsection{Benefits of \sptf}
Since \sptf does not reduce total bytes on wire, and it introduces potential data shuffle overhead and additional communication steps, its advantage may not be obvious instantly. However, \sptf comes with two important properties that significantly benefit training throughput in a typical data center setting: (1) even if \sptf does not reduce bytes on wire, \S\ref{sec:collectives_perf} shows that with the same data volume, running in a smaller world size improves communication throughput. Thus, with a reduction of world size by $L$ times in step f, we expect a handy performance improvement at a large scale; (2) the unique data layout makes it possible to construct dense \textit{tower modules} (detailed later) that can further reduce model complexity and cross-host communication sizes without incurring cross-host synchronization overhead. We evaluate these claims in later sections.

\subsubsection{Specialized \sptf}
\label{sec:specialized_spt}
While we discussed the general case of \sptf, we note that \sptf can be customized to fit different needs. First, a tower can be assigned to a set of $K$ hosts instead of 1, and SPTT will hold without modification when $G \% K = 0$. This allows us to trade off reduction in world size of peer AlltoAll with increased overhead of step c-e. Second, Figure~\ref{fig:tower_transform} showed a general case where embedding tables are assigned as a whole to a GPU, but \sptf can still work in case a table is too large to fit, by leveraging column-wise or row-wise sharding within a tower. For example, for single-hot features, step d is a simple AlltoAll operation that communicates the column-wise shards; but with multi-hot features, we can leverage row-wise sharding and step d becomes a ReduceScatter. Third, depending on the relative data volume of the sparse inputs versus the embedding volume, step b and step c can be swapped so data shuffle on the object with a smaller size. Last but not least, step c can even be omitted by using a virtual process group where the GPUs are already in the peer order.

\begin{figure}[!t]
\begin{center}
	\centering
	\footnotesize
	\includegraphics[width=.8\columnwidth]{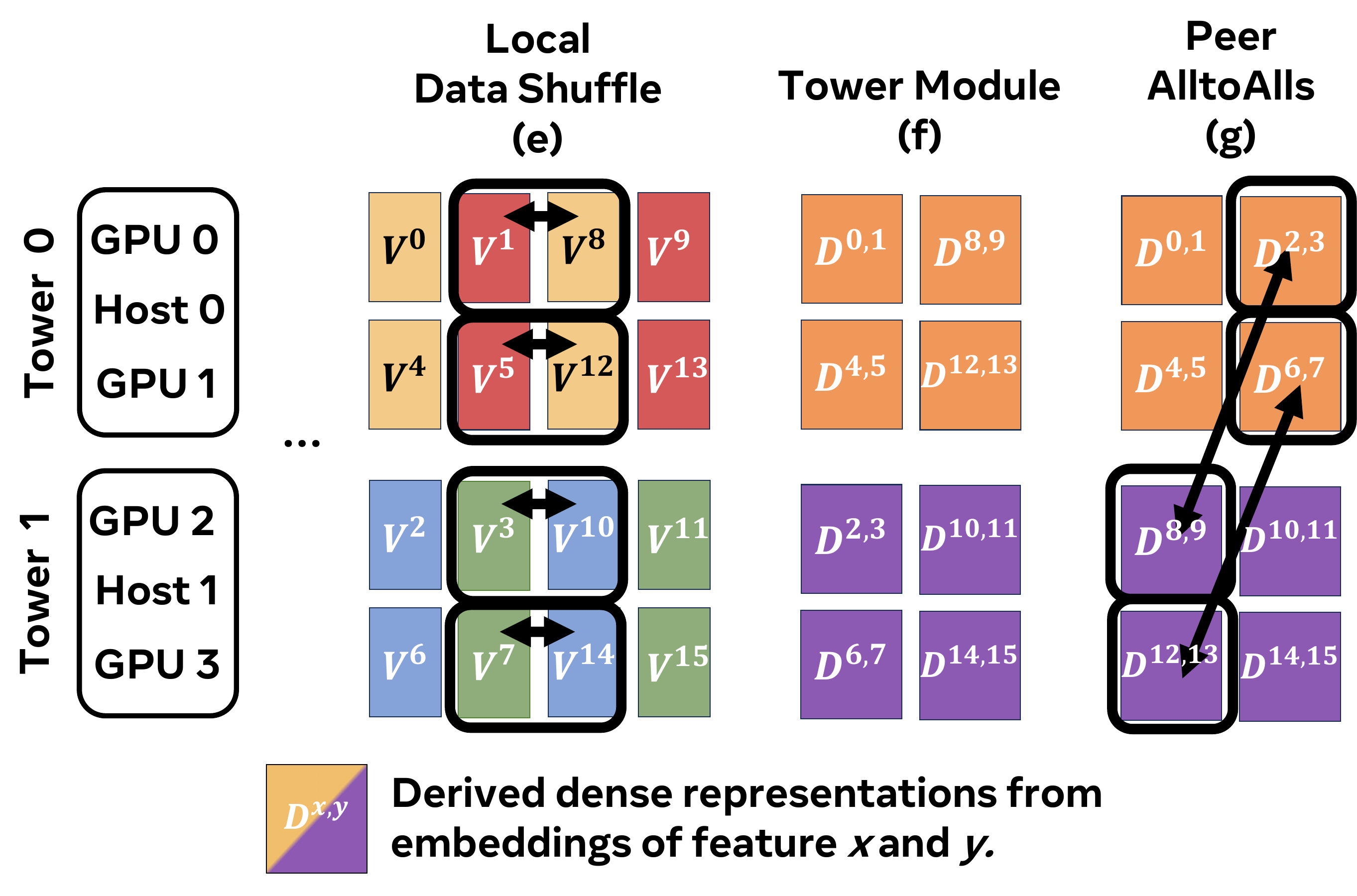}
\end{center}
\vspace*{-3mm}
\caption{Tower Modules and Hierarchical Feature Interaction.}
\label{fig:tower_modules}
\end{figure}

\subsection{Tower Modules and Hierarchical Feature Interaction}
\sptf created a large opportunity to further decrease model complexity and compress cross-host communication volume without incurring large global synchronization overhead or quality degradation through the idea of ~\textit{hierarchical feature interaction}.

To do so, we can construct tower modules (\towermodule) right at the end of step e in \sptf. As shown in Figure~\ref{fig:tower_modules}, we can derive compressed representations for features in a tower which in our setup is $D^{x,y} = TM(V^x, V^y)$, with the expectation that $|D^{x,y}| < |V^x| + |V^y|$ so the network volume at step g can be reduced. We view the benefits of tower modules from two aspects:

\textbf{System Perspective}
\towermodule can further compress output embedding before step f for a reduced cross-host communication volume and computation complexity. \towermodule leverages the fact that data layout of the outcome of \sptf where each GPU holds the embeddings of all features in a tower for all its peers and that GPUs in a tower fully contain the global embeddings for that entire feature group, so the synchronization of \towermodule on the backward pass only needs to happen within a tower (e.g., a host), exploiting the communication locality (e.g., NVLink) for throughput gain. \towermodule can compress the embedding outputs of tables by installing additional dense modules, and its potential reduction of model compute complexity is also sizable: for example, consider a pair-wise dot-product feature interaction module  with $|F|$ features, then the original global feature interaction has a complexity of $O(|F|^2)$; with a dot-product based \towermodule with a feature reduction ratio $r$, the new compute complexity is reduced to $O(\frac{|F|^2}{T^2} + r^2|F|^2)$.

\textbf{Model Perspective}
With potentially reduced output size of each tower produced by \towermodule, the global feature interactions at later stage of the models no longer operate on the full feature set, leading to potential loss in model quality. However, we can compensate for this loss of quality through carefully designed, higher-order feature interactions using \towermodule with the following observations:

\begin{itemize}[leftmargin=*,noitemsep,topsep=0pt]
    \item Feature interaction can be sparse, and hence global feature interaction may have inherent redundancy~\cite{adnan2023ad}. With recent models incorporating multi-domain features~\cite{zhou2023comprehensive,10.1145/3407190}, not all pair-wise feature interaction is meaningful. Thus, by carefully partitioning global features into groups of features with meaningful interaction, we can capture the most important interactions that contribute to model quality with lower computation cost. 
    \item \towermodule essentially introduces an additional order of interaction. Higher order interactions through more stacked interaction layers has been a major contributor to better quality in recommendation models~\cite{wang2021dcn, lian2018xdeepfm, kang2020learning, zhang2024wukong}. \towermodule can be viewed as a special mechanism to achieve higher order interactions as it captures group-level interactions followed by cross-group interactions in a hierarchical manner. Further, with inherent heterogeneity in the \towermodule, the ability to capture a diverse set of interaction is enhanced, which is shown to be more effective than stacking homogeneous interaction modules alone~\cite{dhen}.
    \item Mixture of Experts (MoE) is a powerful paradigm that extends the capacity and expressiveness of models~\cite{fedus2022switch,lepikhin2020gshard}. Traditional application of MoEs is on the batch dimension through a gated network, and on the contrary, \towermodule can be viewed as a MoE on the feature dimension.
\end{itemize}

The following sections detail how we create such meaningful feature partitions and design concrete tower architectures for specific recommendation models. 


\subsection{Learned, Balanced, and Meaningful Feature Partition via Tower Partitioner}
We must take both desired system property (balanced, so that work is balanced across hardware) and model property (cohesive, so that features that have meaningful interactions are kept together) into account when producing feature partitions to generate towers. Traditional approaches to this problem usually involves some form of feature interaction probes followed by a manual process for balanced assignments. This approach can be cumbersome when the natural number of feature clusters do not match the desired tower sizes (data center topology). To solve this issue, we propose a learned, balanced, end-to-end Tower Partitioner (TP). 

Without lose of generality, assume we have a feature interaction result $I$ of dimension ($|F|$, $|F|$), obtained from an original model with normalized feature embedding $F_i$ (a $N$-dimensional embedding vector for feature $i$), using a chosen kernel $K$, where $|F|$ is the cardinality of features in the model. In \dmtembed, we limit our discussion to $K$ being the cosine similarity kernel $K = F_i^TF_j$ due to its popularity. Specifically,

\[I(i,j) = abs(K(F_i, F_j))\]

This yields a interaction matrix with values between 0 and 1, where 0 indicates a pair of features have nothing in common (orthogonal), and 1 indicates the features are strongly positively or negatively related. We then convert $I$ into a distance matrix $D = f(I)$. Notice that this conversion need not preserve the exact distance, as long as it preserves the relative distance among embeddings. 

With $D$, we embed features in a Euclidean space of dimension $n$, obtaining a set of coordinates for each feature $X_i = (x_{i,1}, x_{i,2}, ..., x_{i,n})$ whose pairwise distance corresponds to distance matrix $D$. To save computation, and to reduce noise in the embedding process, we have $n < N$. We use an optimizer (e.g., Adam) to solve for $X_i$ to minimize the following objectives:

\[\sum_{i=1}^{|F|}\sum_{j=1}^{i-1}(d_{i,j} - D(i,j))^2\]

where $d_{i,j} = \|X_i - X_j\|_2$ is the Euclidean distance between embedded coordinates for feature $i$ and $j$.

We then use a constrained K-Means~\cite{bradley2000constrained} algorithm to partition the embedded features into balanced groups, with the constraint that the maximum group size to be at most $K$ (tunable) times the minimum tower size.

One may question why not directly use the raw embedding $F$s for \dmtembed. This is because in a minibatched training setting with batch size $B$, the raw embedding $R$ is of shape $(B,F,N)$, and averaging along $B$ dimension has no meaning because the embedding index for each feature can be different across batches. And instead, opting for  the average feature affinity with $mean(RR^T, dim=0)$ is semantically meaningful because the similarity between features across samples should be coherent.

Compared to traditional graph cut-based, NP-hard algorithms~\cite{Minimumk35:online} which usually optimizes only for an aggregate metrics in each partition (e.g., sum of similarities within each partition) that may not have inherent meaning, \dmtembed provides an efficient means to directly optimize for the target metrics. \dmtembed provides two ways to find meaningful partitions: the first is to assign features that are more different to a group, increasing diversity in each tower, and this is done by setting $f(I) = I$ when obtaining $D$. We call this the \textit{diverse strategy}; the second is to assign features that are more similar to a group, increasing coherence in each tower: this can be achieved by setting $f(I) = 1 - I$ to derive $D$ and we call this the \textit{coherent strategy}. We believe the better choice can vary by model and dataset, and we simply try both to find the optimal setting.

\section{Implementation and Optimization}
\label{sec:implementation}
We built DMT in PyTorch on top of Neo~\cite{mudigere2021high} and TorchRec~\cite{torchrec}. This section describes a few implementation details.

\textbf{Embedding Table Sharding} We leverage the framework's embedding sharder to support placement of embedding tables within a tower to individual GPUs. Specifically, for a large batch size and single-hot embedding, we pin embedding types to column-wise shard due to its lower communication volume, and for small-batch size and multi-hot embedding, row-wise sharding is used.


\textbf{Tower Module Architectures}
We introduce two concrete \towermodule architectures for two widely-used recommendation models by industry, DLRM and DCN. Note our goal here is not to highlight specific module implementations but to demonstrate a systematic and practical way of constructing \towermodule by lifting the main interaction architecture from the underlying model. In both cases, we constrained our choice of operators from the ones used in the interaction arch when building \towermodule. Each \towermodule takes in two inputs: a \textit{embs} of shape $(B, |F|, N)$ which is the output of step e in Figure~\ref{fig:tower_transform} for each tower and a output feature dimension $D$. 

\begin{lstlisting}[language=Python,caption=Tower Module for DLRM, basicstyle=\small\ttfamily,numbers=none]
 def forward(embs, c, p, D):
    B, F, N = embs.shape
    embs_flat = embs.view(B, -1)
    o1 = linear(in_f = N*F, o_f = p*D)(embs_flat).view(B, -1)
    o2 = linear(in_f = N, o_f = c*D)(embs).view(B, -1)
    return cat([o1, o2], dim = 1)
\end{lstlisting}
\label{code:dlrm_tm}

In DLRM, the tower architecture is an ensemble of a simple linear combination of the input features and a projection on the embedding dimension. These operators resemble those used in DLRM overarch. The parameter $c$ and $p$ controls the output feature count for these two linear operators. The output dimension becomes $O = D(c|F| + p)$. 

\begin{lstlisting}[language=Python, caption=Tower Module for DCN,basicstyle=\small\ttfamily,numbers=none]
 def forward(embs, D):
    B, F, N = embs.shape
    o = crossnet(F*N, ...)(embs)
    return linear(in_f = F*N, o_f = F*D)(o).view(B, -1)
\end{lstlisting}

As an optimization, we implement dot product with a manual pair-wise routine when the batch size is large and $|F|$ is relatively small, as we find the generated \textit{cublasGemvTensorStridedBatched} CUDA kernel vastly slower than the manual routine on the data layout we have.

The \towermodule in DCN is a smaller CrossNet, the main interaction module in DCN, with an output dimension of $O = |F|D$.
In both cases, we can derive the \textit{compression ratio} of network communication for step e in Figure~\ref{fig:tower_transform} as $CR = \frac{\sum_{i=0}^TO_i}{|F|N}$.

\section{Evaluation}
\label{sec:evaluation}
We now assess the ability of DMT to generalize across models and preserve model quality, and highlight its throughput gain over state-of-the-art baselines with a detailed ablation study of the factors that contributed to DMT's gain.

\subsection{Experimental Setup}
\textbf{Models} We consider two families of recommendation models. The first family is the open-source models which include two types of interactions: dot-product (DLRM) and CrossNet (DCN). They share the same embedding table sizes and have about 90GB of total parameters and consumes 14 to 96MFlops/sample to train. The second family contains an internal, extra-large model (\lrm for short) similar to those described in industry~\cite{mudigere2021high,lian2021persia}. This model contains about 2 trillion parameters and has a complexity around 700~MFlops/sample.

To create meaningful and balanced partitions for DMT, we use the dot-product based \dmtembed on a 2D plane with $R=1$ for constrained K-Means. We use this grouping for both training throughput and accuracy evaluation. We pin each tower module to a single host to best leverage NVLink. 

\textbf{Hardware}
Where applicable, we run our experiments on multiple hardware platforms based on the Nvidia V100, A100 and H100 GPUs. We vary the number of GPUs used in our benchmarks from 16 to 512. Our infrastructure guarantees full bisection bandwidth between any pair of hosts with no oversubscription.

\begin{table}[t!h!]
	\centering
	\resizebox{.9\columnwidth}{!}
	{
    	\begin{tabular}{|c|c|c|c|}
            \hline
                Config & Batch Size & AUC & Epoch Time \\
    	\hline
    	    Baseline (DLRM) & 2048 & 0.8030 & 6.5hrs \\
            \hline
                \strongbaseline (DLRM) & 128K &\textbf{ 0.8047} & \textbf{29mins} \\
            \hline
            \hline
                Baseline (DCN) & 128K & 0.7963 & 58mins \\
            \hline
                \strongbaseline (DCN) & 128K &\textbf{ 0.8002} & \textbf{27mins} \\
            \hline
    	\end{tabular}
	}
	\caption{We use a \strongbaseline that achieves better evaluation AUCs and higher throughputs compared to TorchRec's results using the same hardware. Baseline's epoch time is as reported.}
    \label{table:strongbaseline_auc}
\end{table}

\textbf{Strong Baseline} We use the official implementations in the state-of-the-art recommendation framework TorchRec~\cite{dlrmtorc14:online, torchrec} from Meta for the open source models as the baseline. We turn on all applicable optimizations including quantized embedding and gradient communication, pipelined data fetching, and TorchRec's auto planner to determine the best sharding strategy (e.g., column-wise, row-wise, or hierarchical sharding for embedding tables). We exclude data parallelism for embedding tables in the search space because synchronizing embedding tables that way is generally impractical~\cite{mudigere2021high}. To help improve performance, we manually include a column-wise sharding factor to balance system loads when there are more GPUs than embedding tables so TorchRec can tap into the collective bandwidth of the whole cluster. For dense component sharding, we use data parallel, suggested by Alpa, as the optimal strategy (Figure~\ref{fig:alpa_dlrm}). We then build DMT on top. Additionally, we improve on the evaluation accuracy reported by the official implementation, by 0.17\% and 0.39\% respectively for the open source models using the Adam optimizer attached with a tuned learning rate schedule. With these changes incorporated, we refer to the baseline used in the following evaluation as \strongbaseline, whose characteristics are shown in Table ~\ref{table:strongbaseline_auc}.


\subsection{Accuracy and Generalizability}
\label{sec:auc}
We first evaluate \sptf and \towermodule in their generalized abilities to capture interactions and minimize accuracy losses across recommendation models. We summarize model size and complexity as reported by the PyTorch profiler. To properly reflect run to run variance, we run each experiment at least 9 times and report the 1-epoch median evaluation AUC along with its standard deviation. We use the Adam optimizer and keep all hyperparameters identical for DMT and all open source model's experiments for fair comparison. We set global batch sizes to 128K for all models. To test the generalization of DMT as a modeling technique itself, we do not put effort in hyperparameter tuning for individual DMT models, inline with prior art~\cite{narang2021transformer}. We evaluate DLRM and \lrm for 4B and 35B samples on the Criteo and an internal dataset, respectively. 

\begin{table}[t!h!]
	\centering
	\resizebox{.9\columnwidth}{!}
	{
    	\begin{tabular}{|c|c|c|c|}
    	\hline
    	    Model & AUC  (Std) & MFlops/Sample & Parameters (G) \\
    	    \hline
    	    DLRM & 0.8047 (0.0004) & 14.74 & 22.78\\ 
    	    \hline
    	    \sptf-DLRM & 0.8053 (0.0004) & 14.74 & 22.78 \\ 
                \hline
    	    \hline
    	    DCN    & 0.8002 (0.0003) &   96.22    & 22.79 \\
    	    \hline
    	    \sptf-DCN   & 0.8001 (0.0002) & 96.22 & 22.79 \\    	    
    	\hline
    	\end{tabular}
	}
	\caption{\spt achieves neutral AUC compared to unmodified DLRM models.}
    \label{table:spt_auc}
\end{table}

\subsubsection{\titlecap{\spt}}
\label{sec:eval_spt}
We create a separate a pass-through tower for each embedding feature in DMT and compare AUC with the original models to show that since \spt only orchestrates dataflow but leaves compute intact, it has no implication on model quality, as shown in Table~\ref{table:spt_auc}.

\subsubsection{Tower Modules}
We use the \towermodule detailed in \S\ref{sec:implementation} for the open source models. Since \towermodule modifies semantics, we show that we can recover global interactions through hierarchical interaction, provided that there is a sufficient amount of towers. 

\begin{table}[t!h!]
	\centering
	\resizebox{\columnwidth}{!}
	{
    	\begin{tabular}{|c|c|c|c|}
    	\hline
    	    Model & AUC  (Std) & MFlops/Sample & Parameters (G) \\
    	    \hline
    	    DLRM \strongbaseline & 0.8047 (0.0004) & 14.74 & 22.78\\ 
    	    \hline 
    	    DMT 2-DLRM & 0.8046 (0.0004) & 8.95 & 22.78 \\ 
    	    \hline
    	    DMT 4-DLRM & 0.8045 (0.0003) & 8.95 & 22.78 \\ 
    	    \hline 
    	    DMT 8-DLRM & 0.8045 (0.0001) & 8.95 & 22.78 \\ 
    	    \hline
    	    DMT 16-DLRM & 0.8047 (0.0003) & 8.75 & 22.78 \\ 
                \hline
                DMT 26T-DLRM & 0.8047 (0.0004) & 8.95 & 22.78 \\ 
    	    \hline
                \hline
    	    DCN \strongbaseline  & 0.8002 (0.0003) &   96.22    & 22.79 \\
    	    \hline
    	    DMT 2T-DCN   & 0.7998 (0.0002) &  43.71     & 22.81 \\
    	    \hline
    	    DMT 4T-DCN    & 0.8003 (0.0002) & 50.01     & 22.79 \\
    	    \hline
    	    DMT 8T-DCN    & \textbf{0.8006} (0.0002) &   62.60    & 22.81 \\
    	    \hline
    	    DMT 16T-DCN   & 0.8001 (0.0002) &  87.19     & 22.79 \\
                \hline
                DMT 26T-DCN   & 0.8001 (0.0002)                & 96.22  & 22.79     \\
    	\hline
    	\end{tabular}
	}
	\caption{Median evaluation AUC (with standard deviation) comparing open source models with their DMT counterparts, which achieve on par or better AUC with on par or lower resources.}
    \label{table:dlrm_auc}
\end{table}

\textbf{Open Source}  We show the AUC results of applying tower modules on top of SPTT. For DLRM, we set $p=1, c=0, D=128$ for 16 towers, and $c=1, p=0, D=64$ for 2-8, 26 towers. For DCN, we set $D=128$ for 2-16 towers (for 26-tower DCN, we simply use \sptf alone). The baselines use $N = 128$ as the default embedding dimension. While there are other parameters to \towermodule that achieve similar results, we picked these because they achieved best throughput (discussed in the next section). 

It is worth noting that changing the number of towers have impacts on the total parameter count: more towers may introduce more parameters in the tower modules, but can reduce parameters in the over arch. This effect is more pronounced in DCN than DLRM, because dot-product is parameter-free but CrossNet is not. Since the Criteo dataset has 26 sparse features, we set number of towers to 26\footnote{Note that this is not a hard limitation for DMT: if we wish to have more towers to accommodate to a larger cluster size, we can always perform column sharding to embedding tables.}. 

Table~\ref{table:dlrm_auc} summarizes the results: DMT is able to achieve on par (within one standard deviation) or better AUC (more than one standard deviation) with on par parameter count and flops usage. 

\begin{table}[t!h!]
    \centering
    \resizebox{.9\columnwidth}{!}
    {
        \begin{tabular}{|c|c|c|c|}
        \hline
            Compression Ratio & AUC  (Std) & MFlops/Sample & Parameters (G) \\
            \hline
            2 & 0.8045 (0.0001) & 8.95 & 22.78 \\ 
            \hline
            4 & 0.8036 (0.0003) & 7.93 & 22.78 \\
            \hline
            8 & 0.8022 (0.0004) & 7.41 & 22.78 \\
            \hline
            16 & 0.8000 (0.0003) & 7.16 & 22.78 \\
            \hline
        \end{tabular}
    }
    \caption{Median evaluation AUC (with standard deviation) versus varying tower compression ratio of DMT 8T-DLRM.}
    \label{table:dlrm_compressopn_ratio}
\end{table}

To assess how AUC varies according to the compression ratio ($CR$), we set $D$ of towers in DMT 8T-DLRM to 64, 32, 16 and 8, with a corresponding compression ratio of 2, 4, 8 and 16. We summarize the result in Table~\ref{table:dlrm_compressopn_ratio}: we observe an expected gradual degradation of AUC with larger compression ratio used in the tower arch.

\textbf{\lrm}
\label{sec:lrm_auc}
We partition the sparse features into 16 towers and keep the tower module's operators same as its main interaction type. Since the \lrm's embedding table sizes are significantly larger than those in the open source models, we use a column-wise sharding scheme to slice embedding tables into equal-dimension chunks, and invoke \dmtembed for tower partition. We train both the original model and the DMT model on a CTR task, and we observe a normalized entropy~\cite{10.1145/2648584.2648589} improvement of 0.02\%.

\subsubsection{Effectiveness of \dmtembed}
We now assess the role of \dmtembed in creating meaningful feature partitions and how it affects model quality. 

\begin{figure}[!t]
\begin{center}
	\centering
	\footnotesize
	\includegraphics[width=\columnwidth]{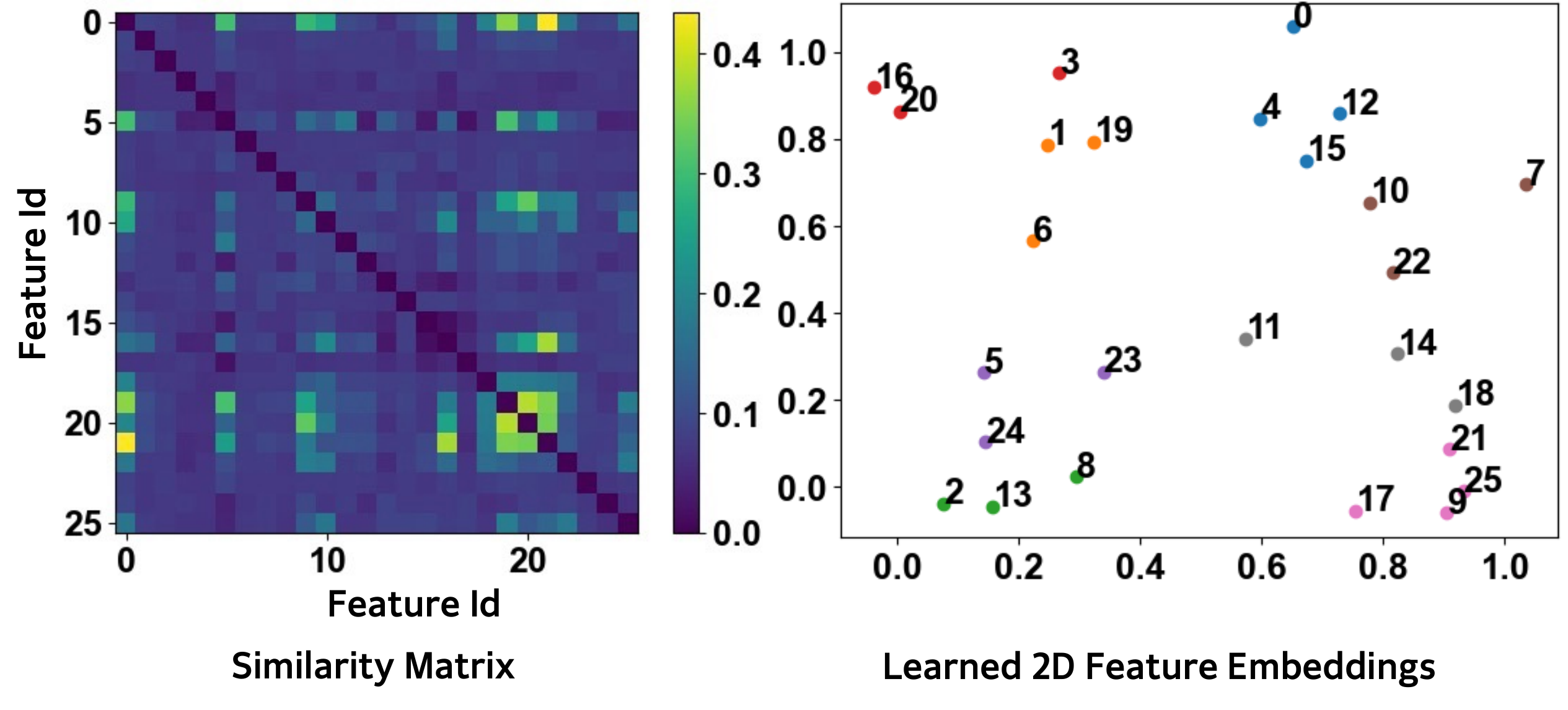}
\end{center}
\vspace*{-3mm}
\caption{The similarity matrix of \dmtembed using the \textit{coherent} strategy and partition of features into 8 color-coded towers.} 
\label{fig:dmt_embed}
\end{figure}

\begin{table}[t!h!]
    \centering
    \resizebox{.9\columnwidth}{!}
    {
        \begin{tabular}{|c|c|c|c|}
        \hline
            Config (LR) & \dmtembed (Std) & Naive (Std) & p-Value\\
            \hline
            DMT 16T-DLRM (1e-3) & \textbf{0.7990} (0.0003) & 0.7981 (0.0003)  & 0.0006  \\ 
            \hline
            DMT 8T-DCN  (2e-3) & \textbf{0.8006} (0.0002) & 0.8003 (0.0003) & 0.0023 \\

        \hline
        \end{tabular}
    }
    \caption{\dmtembed creates more meaningful feature to tower assignment compared to a naive assignment, achieving better median AUC with statistical significance in various settings.}
    \label{table:auc_dmt_embed}
\end{table}

\textbf{Open Source} To demonstrate how \dmtembed improves AUC under various hyperparameter settings, we introduce a balanced but naive, sequential feature to tower assignment scheme with a stride equal to the number of towers as the baseline, e.g., for 8 towers the resulting assignment is [[0, 8, 16, 24], [1, 9, 17, 25], [2, 10, 18], [3, 11, 19], [4, 12, 20], [5, 13, 21], [6, 14, 22], [7, 15, 23]]. We report the resulting AUC using \dmtembed and this assignment in Table~\ref{table:auc_dmt_embed}. Evidently, \dmtembed's partition leads to higher model quality. We derive the staticstical signifcance of our results by performing the Mann-Whitney U test for 9 repeated experiments. With p values low enough we reject the null hypothesis that two experiments using \dmtembed and naive assignments have equal chance of yielding better AUC in these settings.


\textbf{\lrm} We find patterns in $D$ captured by \dmtembed mostly manifest as interactions between dedicated item, item-user, and dedicated user features. This naturally leads us to first partition features into these 3 categories, and further partition them through \dmtembed into individual towers of desirable counts. Our experiments suggests this strategy of partition is the main contributor of the NE improvement observed in \S\ref{sec:lrm_auc}. Our further experiments suggest once we have features grouped into these three groups, exact partitions of these groups into desired towers have marginal effect. 

\subsection{Performance and Scalability}
This section demonstrates the throughput improvement of DMT across a wide range of hardware platforms and scales. To minimize variance introduced by the data ingestion pipeline, we use a random dataset for throughput evaluation.

\subsubsection{Overall Throughput Gain}
We first look at the overall speedup of DMT. 

\begin{figure*}[!htb]
\begin{center}
	\centering
	\footnotesize
	\includegraphics[width=\linewidth]{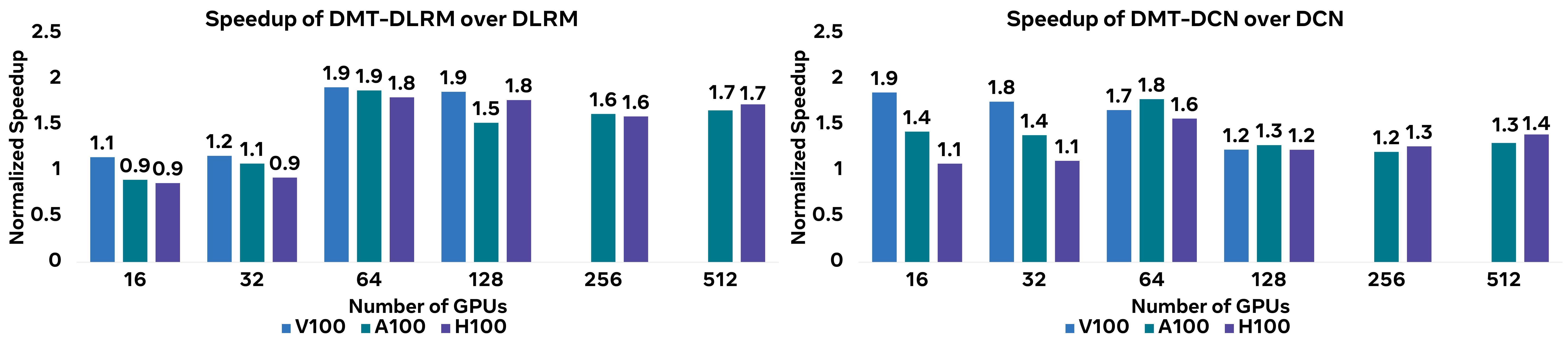}
\end{center}
\vspace*{-5mm}
\caption{Speedup of DMT over DLRM and DCN across different hardware platforms and scales. } 
\label{fig:perf_scalability}
\end{figure*}

\textbf{Open Source}
We vary the number of towers in DMT to match the number of hosts from 16 to 64. Each partition corresponds to one configuration in \S\ref{sec:auc}. We keep the local batch size fixed in this experiment at 16K.

We summarize the result in Figure~\ref{fig:perf_scalability}. Overall, DMT is able to achieve up to 1.9X and 1.8X speedup across 3 hardware platforms\footnote{Since our V100 cluster can support at most 16 hosts, no data is provided for the 32, 64-host cases.} at a large scale. On the DLRM side, we see a general trend of increasing speedup as we upscale our cluster, because the effects of the reduced communication volume from tower modules outweigh the overhead of data shuffle and compute latency from \sptf and tower modules as scale grows. On the DCN side, DMT scores a large speedup at a small scale. This is mostly due to the reduced model complexity of DMT as characterized in Table~\ref{table:dlrm_auc}. This phenomena is especially pronounced in older GPUs, which is more bounded by compute capacity. 




\textbf{\lrm}
We train DMT and the original \lrm on 128 V100 and A100 GPUs. DMT-\lrm is Note that DMT-XLRM achieves a lower speedup compared to open source models because XLRM is much more compute-bound.

\subsubsection{Ablation Study}
We conduct a comprehensive ablation study on the open source models to understand the origin of speedups by analyzing the effect of \sptf and \towermodule on throughput and quantifying how component latency has improved with DMT.

\begin{figure}[t]
\begin{center}
    \centering
    \footnotesize
    \includegraphics[width=.98\linewidth]{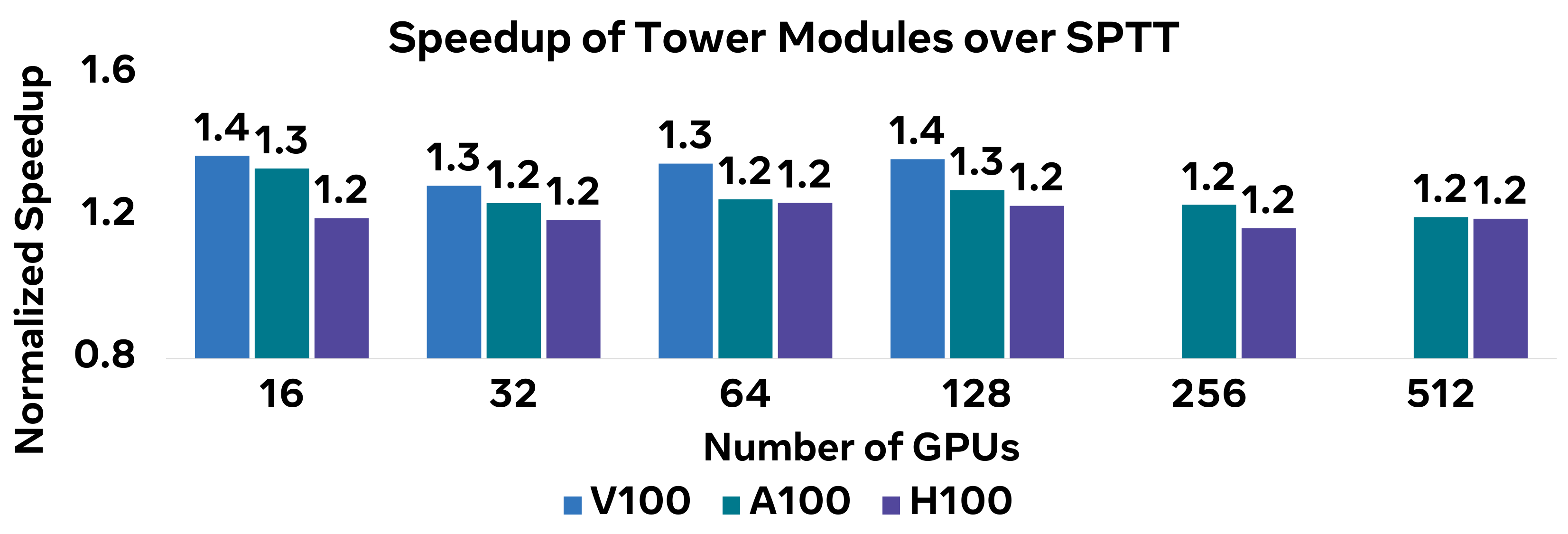}
\end{center}
\vspace*{-3mm}
\caption{Speedup of tower modules over \sptf on DLRM.} 
\label{fig:ablation_tower_spt}
\end{figure}

\begin{figure}[t]
\begin{center}
    \centering
    \footnotesize
    \includegraphics[width=.95\linewidth]{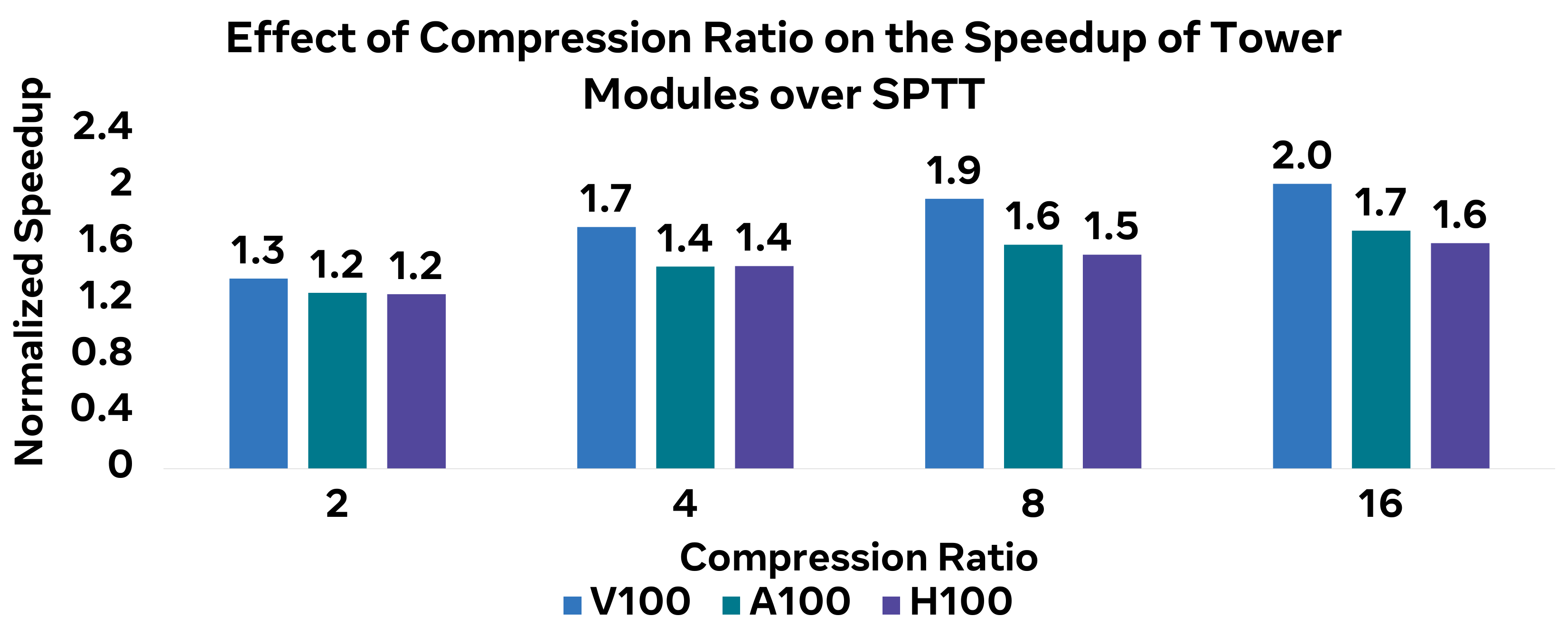}
\end{center}
\vspace*{-3mm}
\caption{Effects of compression ratio on the speedup of DMT 8T-DLRM over DMT-\sptf.} 
\label{fig:compression_ratio}
\end{figure}

\textbf{Effects of \sptf and \towermodule}
We compute the speedup of DMT with \towermodule over DMT \sptf-only from 16 to 512 GPUs (2-26 towers using the same configuration as in Figure~\ref{fig:perf_scalability}), and summarize in Figure~\ref{fig:ablation_tower_spt}. Noticeably, as the training scale grows, the effects of tower modules grow, contributing up to 1.4X additional throughput gain. 

\textbf{Effects of Compression Ratio in \towermodule}
Compression ratio provides a flexible mechanism to tradeoff AUC with throughput. Figure~\ref{fig:compression_ratio} quantify this on DMT 8T-DLRM. Unsurprisingly, by cross-referencing with Table~\ref{table:dlrm_compressopn_ratio}, we observe an increased throughput of up to 2X at a cost of AUC of $<0.5\%$ with a high compression ratio of 16.

\begin{figure}[!t]
\begin{center}
	\centering
	\footnotesize
	\includegraphics[width=\columnwidth]{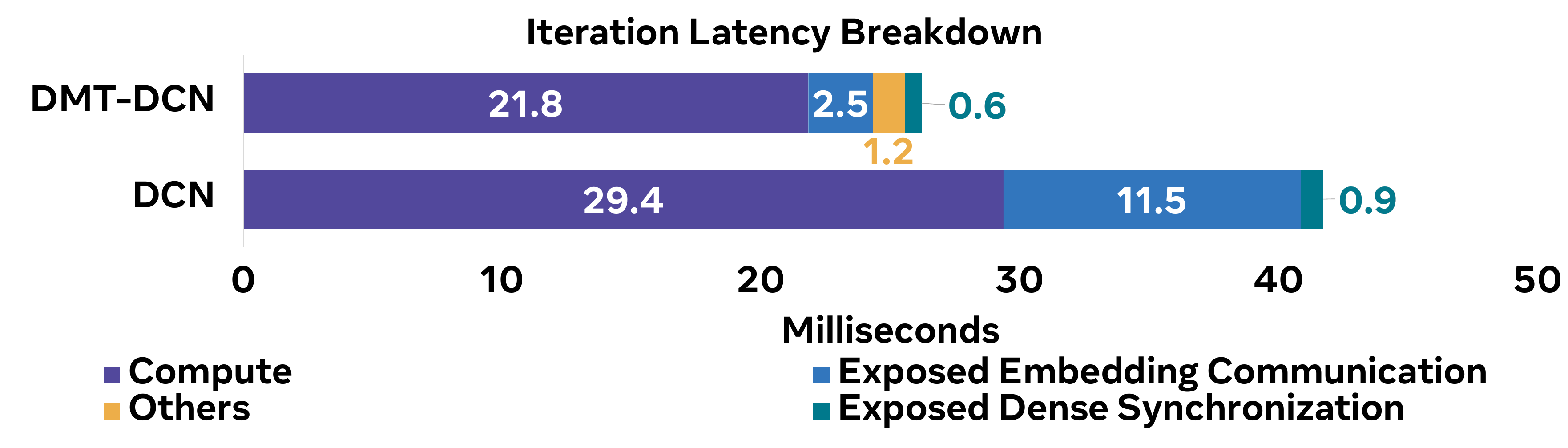}
\end{center}
\vspace*{-3mm}
\caption{DMT improves training latency of all components.}
\label{fig:dmt_lat_breakdown}
\end{figure}

\textbf{Reduction in Component Latency}
Finally, we provide a breakdown of iteration latency comparing DMT-DCN and DCN to quantify the reduction of latency in each component in Figure~\ref{fig:dmt_lat_breakdown}, running on 64 H100 GPUs. Evidently, DMT-DCN sees most improvements in latency from compute (1.4X) and exposed embedding communication (4.6X). 
\section{Discussion}
\label{sec:discussion}

\textbf{Hierarchical Collectives and \sptf} The use of hierarchical collectives for improving performance is not new, as explored by MICs~\cite{zhang2022mics}, FSDP~\cite{zhao2022communication}, DeepSpeed-MoE~\cite{rajbhandari2022deepspeed}, PLink~\cite{luo2020plink}, NetHint~\cite{chen2022nethint}, Heirring~\cite{10.5555/3433701.3433759} and PXN~\cite{Doubling9:online}. However, the novelty of \sptf is as follows: (1) it is an unique application of hierarchical collectives to break down the monolithic embedding lookup process which can be specialized to accelerate different pooling types using highly efficient local shuffles and communications (\S\ref{sec:specialized_spt}); (2) while hierarchical collectives does not reduce bytes on wire, the synergy between \sptf and tower modules make it natural to further compress cross-host communication volumes, making significant additional speedups possible (\S\ref{sec:eval_spt}); (3) compared to hierarchical collectives, \sptf incurs overhead per iteration instead of per collectives.



\textbf{Quantization} Compared to quantization, DMT may result in better model quality and provide asymptotically better speedup. Our extended experiment shows quantizing the XLRM model to FP8 already causes 0.1\% significant quality degradation without extensive tuning; further, on 1024 H100 GPUs, quantized DMT-XLRM can still outperform FP8-quantized XLRMs by up to 1.2X in throughput. 


\section{Conclusion}
\label{sec:conclusion}
We proposed DMT, a novel topology-aware modeling technique for recommendation models that improves throughput by up to 1.9$\times$ at a large scale ($>$ 64 GPUs) while preserving model quality. DMT achieved this through \spt, a novel training paradigm that exploits data center locality, tower modules, synergistic modules that reduces and compresses model complexity and communication volumes, and \dmtembed, a learned, balance feature partitioner that creates meaningful towers. 

\bibliography{mlsys_ver}
\bibliographystyle{mlsys2024}

\end{document}